# Pathologist-Level Grading of Prostate Biopsies with Artificial Intelligence

Peter Ström, M.Sc.[1*], Kimmo Kartasalo, M.Sc.[2*], Henrik Olsson, M.Sc.[1], Leslie Solorzano, M.Sc.[3], Brett Delahunt, M.D.[4], Daniel M Berney, M.D.[5], David G Bostwick, M.D.[6], Andrew J. Evans, M.D.[7], David J Grignon, M.D.[8], Peter A Humphrey, M.D.[9], Kenneth A Iczkowski, M.D.[10], James G Kench, M.D.[11], Glen Kristiansen, M.D.[12], Theodorus H van der Kwast, M.D.[7], Katia RM Leite, M.D.[13], Jesse K McKenney, M.D.[14], Jon Oxley, M.D.[15], Chin-Chen Pan, M.D.[16], Hemamali Samaratunga, M.D.[17], John R Srigley, M.D.[18], Hiroyuki Takahashi, M.D.[19], Toyonori Tsuzuki, M.D.[20], Murali Varma, M.D.[21], Ming Zhou, M.D.[22], Johan Lindberg, Ph.D[1], Cecilia Bergström, Ph.D [23], Pekka Ruusuvuori, Ph.D [2], Carolina Wählby, Ph.D [3,24], Henrik Grönberg, M.D.[1,25], Mattias Rantalainen, Ph.D [1], Lars Egevad, M.D.[26], and Martin Eklund, Ph.D [1]

*Both authors contributed equally to this study.*

Corresponding author: Dr. Martin Eklund; Department of Medical Epidemiology and Biostatistics, Karolinska Institutet, PO Box 281, SE-171 77 Stockholm, Sweden; martin.eklund@ki.se; +46 737121611

1. Department of Medical Epidemiology and Biostatistics, Karolinska Institutet, Stockholm, Sweden.
2. Faculty of Medicine and Health Technology, Tampere University, Tampere, Finland.
3. Centre for Image Analysis, Dept. of Information Technology, Uppsala University, Uppsala, Sweden.
4. Department of Pathology and Molecular Medicine, Wellington School of Medicine and Health Sciences, University of Otago, Wellington, New Zealand.
5. Barts Cancer Institute, Queen Mary University of London, London, UK.
6. Bostwick Laboratories, Orlando, FL, USA.
7. Laboratory Medicine Program, University Health Network, Toronto General Hospital, Toronto, ON, Canada.
8. Department of Pathology and Laboratory Medicine, Indiana University School of Medicine, Indianapolis, IN, USA.
9. Department of Pathology, Yale University School of Medicine, New Haven, CT, USA.
10. Department of Pathology, Medical College of Wisconsin, Milwaukee, WI, USA.
11. Department of Tissue Pathology and Diagnostic Oncology, Royal Prince Alfred Hospital and Central Clinical School, University of Sydney, Sydney, NSW, Australia.
12. Institute of Pathology, University Hospital Bonn, Bonn, Germany.
13. Department of Urology, Laboratory of Medical Research, University of São Paulo Medical School, São Paulo, Brazil.
14. Pathology and Laboratory Medicine Institute, Cleveland Clinic, Cleveland, OH, USA.
15. Department of Cellular Pathology, Southmead Hospital, Bristol, UK.
16. Department of Pathology, Taipei Veterans General Hospital, Taipei, Taiwan.
17. Aquesta Uropathology and University of Queensland, Brisbane, Qld, Australia.
18. Department of Laboratory Medicine and Pathobiology, University of Toronto, Toronto, ON, Canada.
19. Department of Pathology, Jikei University School of Medicine, Tokyo, Japan.
20. Department of Surgical Pathology, School of Medicine, Aichi Medical University, Nagoya, Japan.
21. Department of Cellular Pathology, University Hospital of Wales, Cardiff, UK.
22. Department of Pathology, UT Southwestern Medical Center, Dallas, TX, USA.
23. Department of Immunology, Genetics and Pathology, Uppsala University, Uppsala, Sweden.
24. BioImage Informatics Facility of SciLifeLab, Uppsala, Sweden.
25. Department of Oncology, S:t Göran Hospital, Stockholm, Sweden.
26. Department of Oncology and Pathology, Karolinska Institutet, Stockholm, Sweden.

# Abstract

**Background:** An increasing volume of prostate biopsies and a world-wide shortage of uro-pathologists puts a strain on pathology departments. Additionally, the high intra- and inter-observer variability in grading can result in over- and undertreatment of prostate cancer. Artificial intelligence (AI) methods may alleviate these problems by assisting the pathologist to reduce workload and harmonize grading.

**Methods:** We digitized 6,682 needle biopsies from 976 participants in the population based STHLM3 diagnostic study to train deep neural networks for assessing prostate biopsies. The networks were evaluated by predicting the presence, extent, and Gleason grade of malignant tissue for an independent test set comprising 1,631 biopsies from 245 men. We additionally evaluated grading performance on 87 biopsies individually graded by 23 experienced urological pathologists from the International Society of Urological Pathology. We assessed discriminatory performance by receiver operating characteristics (ROC) and tumor extent predictions by correlating predicted millimeter cancer length against measurements by the reporting pathologist. We quantified the concordance between grades assigned by the AI and the expert urological pathologists using Cohen's kappa.

**Results:** The AI achieved an area under the ROC curve of 0.997 for distinguishing between benign and malignant biopsy cores, and 0.999 for distinguishing between men with or without prostate cancer. The correlation between millimeter cancer predicted by the AI and assigned by the reporting pathologist was 0.96. For assigning Gleason grades, the AI achieved an average pairwise kappa of 0.62. This was within the range of the corresponding values for the expert pathologists (0.60 to 0.73).

**Conclusions:** The performance of the AI to detect and grade cancer in prostate needle biopsy samples was comparable to that of international experts in prostate pathology. AI has potential to reduce high intra-observer variability and to provide diagnostic expertise in regions where this is currently unavailable.

# Introduction

Histopathological evaluation of prostate biopsies is critical to the clinical management of men suspected of having prostate cancer. Despite this importance, the histopathological diagnosis of prostate cancer is associated with several challenges:

- More than one million men undergo prostate biopsy in the United States annually.[1] With the standard biopsy procedure resulting in 10-12 needle cores per patient, this means that more than 10 million tissue samples need to be examined by pathologists. The increasing incidence of prostate cancer in an aging population means that the number of biopsies is likely to further increase.
- It is recognized that there is a shortage of pathologists internationally. In China, there is only one pathologist per 130,000 population, while in many African countries the ratio is of the order of one per million.[2,3] Western countries are facing similar problems, with an expected decline in the number of practicing pathologists due to retirement.[4]
- Gleason grade is the most important prognostic factor for prostate cancer and is crucial for treatment decisions. Gleason grade is based on morphologic examination and is recognized to be notoriously subjective. This is reflected in high intra- and inter-pathologist variability in reported grades, as well as both under- and over-diagnosis of prostate cancer.[5,6]

A possible solution to these challenges is the application of artificial intelligence (AI) to prostate cancer histopathology. The development of an AI to identify benign biopsies with high accuracy would decrease the workload of pathologists and allow them to focus on difficult cases. Further, an accurate AI could assist the pathologist with the identification, localization and grading of prostate cancer among those biopsies not culled in the initial screening process, thus providing a safety net to protect against potential misclassification of biopsies. AI-assisted pathology assessment could harmonize grading and reduce inter-observer variability, leading to more consistent and reliable diagnoses and better treatment decisions.

Using high resolution scanning, tissue samples can be digitized to whole slide images (WSI) and utilized as input for the training of deep neural networks (DNN), an AI technique which has been successful in many fields, including medical imaging.[7–10] Despite the many successes of AI, little work has been undertaken in prostate diagnostic histopathology.[11–16]

Attempts at grading prostate biopsies by DNNs have been limited to small datasets or subsets of Gleason patterns, and they have lacked analyses of the clinical implications of the introduction of AI-assisted prostate pathology.

In this study, we aimed to develop an AI with clinically acceptable accuracy for prostate cancer detection, localization, and Gleason grading. To achieve this, we digitized 8,313 samples from 1,222 men included in the prospective and population based STHLM3 prostate cancer diagnostic study undertaken in 2012-2015.[17,18] We evaluated the performance of the model on an independent test set and through a comparison with 87 cases of prostate cancer graded by the International Society of Urological Pathology (ISUP) Imagebase panel consisting of 23 experienced uro-pathologists.[19]

# Methods

## Sample population and data collection

Between 2012 and 2015, the prospective and population-based STHLM3 study evaluated a diagnostic model for prostate cancer in men aged between 50 and 69 years.[17,18] Among the 59,159 participants, 7,406 (12.5%) underwent systematic biopsy according to a standardized protocol consisting of 10 or 12 needle cores; with 12 cores being taken from prostates larger than 35 cm$^3$ (Table 1). Urologists who participated in the study and the study pathologist were blinded to the clinical characteristics of the patients. A single pathologist (L.E.) graded all biopsy cores according to the ISUP grading classification (where Gleason scores 6, 3+4=7, 4+3=7, 8, and 9-10 are reported as ISUP grade 1 to 5, also referred to as Gleason Grade Groups) and delineated cancerous areas using a marker pen.[20,21]

The biopsy cores were formalin fixed and stained with hematoxylin and eosin. A selection of 8,313 biopsies from 1,222 STHLM3 participants was digitized. The cases were chosen to represent the full range of diagnoses, with an over-representation of high-grade disease. We used images from 1,631 cores from a random selection of 246 (20%) men to evaluate the performance of the AI, while the rest were used for model training. That is, *all* biopsies from a given man were assigned to either the training or the test dataset.[22] In addition, to further enrich the data, 152 slides from men with prostate cancers of the highest grades (ISUP 4 and 5) were collected from Capio S:t Göran Hospital, Stockholm. These slides were re-graded by L.E. and utilized for training purposes only. As an additional test set, we digitized 87 cores from the Pathology Imagebase, a reference database launched by ISUP.[19] Test images, as well as all cores from Imagebase patients, were not part of model development and were excluded from any analysis until the final evaluation. For details concerning data collection, see Appendix.

## Artificial intelligence framework

### Image pre-processing

We processed the WSIs with a segmentation algorithm based on Laplacian filtering to identify the regions corresponding to tissue sections and annotations drawn adjacent to the tissue (Figure S1). We then extracted digital pixel-wise annotations, indicating the locations of cancerous tissue of any grade, by identifying the tissue region corresponding to each annotation. To obtain training data representing the morphological characteristics of Gleason patterns 3, 4 and 5, we extracted numerous partially overlapping smaller images, or *patches*, from each WSI. Each patch was small enough to largely represent only benign or cancerous tissue. We used patch dimensions of 598 x 598 pixels (approx. 540 x 540 µm) at a resolution corresponding to 10X magnification (pixel size approx. 0.90 µm). The process resulted in approximately 5.1 million patches usable for training a DNN. See Appendix for details (Table S1).

### Deep neural network model for classification of image patches

We used two DNN ensembles, each consisting of 30 Inception V3 models pre-trained on ImageNet, with classification layers adapted to our outcome.[23,24] The first ensemble performed binary classification of image patches into benign or malignant, while the second ensemble classified patches into Gleason patterns 3 to 5. To reduce label noise in the latter case, we trained the ensemble on patches extracted from cores containing only one Gleason pattern (i.e. cores with Gleason score 3+3, 4+4, or 5+5). Importantly, the test data still contained cores of all grades to provide a real-world scenario for evaluation. Each DNN in the first and the second ensemble thus predicted the probability of each patch being malignant, and whether it represented Gleason pattern 3, 4, or 5, respectively. See Appendix for details (Figure S2).

### Boosted tree model for core-level estimation of cancer grade and length

Once the probabilities for the Gleason pattern at each location of the biopsy core were obtained from the DNN ensembles, we mapped them to core-specific characteristics (ISUP grade and cancer length) using boosted trees.[25] All cores in the training data were used for training the boosted trees. Specifically, aggregated features from the patch-wise probabilities predicted by each DNN for each core were used as input to the boosted trees, and the clinical assessment of ISUP score and cancer length were used as outcomes. See Appendix for details.

# Evaluation metrics

## Cancer detection

We summarized the operating characteristics of the AI system in a Receiver Operating Characteristic (ROC) curve and the Area Under the ROC Curve (AUC). We then specified a range of acceptable sensitivities for potential clinical use, and evaluated achieved specificity (both on core-level and patient-level) when compared to the pathology report.

## Cancer length estimation

We predicted cancer length in each core and compared it to the cancer length described in the pathology report. The comparison was undertaken on individual cores as well as on aggregated cores (i.e. total cancer length) for each man. Linear correlation was assessed on both all cores and men, as well as restricted to positive cores and men. In addition, the patch-wise predictions were mapped to their spatial locations on each biopsy core, thus permitting the generation of visualizations of the predictions.

## ISUP grading

Cohen's kappa with linear weights was used for evaluating the AI's performance against pathologists on the Imagebase test set. Linear weights emphasize a higher level of disagreement of ratings further away from each other on the ordinal ISUP scale, in accordance with previous publications on the Imagebase study.[19] Each of the 87 slides in Imagebase was graded by each of the 23 Imagebase panel pathologists, and additionally by the AI. To evaluate how well the AI agreed with the pathologists, we calculated all pair-wise kappas and summarized the average for each of the 23 raters. In addition, we estimated the kappa with a grouping of the Gleason scores in ISUP grades (grade groups) 1, 2-3 and 4-5.

# Results

## Cancer detection

We estimated the AUC representing the ability of the AI to distinguish malignant from benign cores to 0.997 (Figure 1). For predicting whether a man had cancer or not, the AUC was 0.999. As an example (Figure 1; second row from top), at a sensitivity of 99.6%, the AI achieved a specificity of 86.6%. At this sensitivity level, the AI failed to detect three cores with cancer (two ISUP grade 1 and one ISUP grade 2, all with less than 0.5 mm cancer) across 721 malignant biopsy cores in the independent test data. No cancer was misdiagnosed since other malignant cores from the same men were correctly classified.

## Cancer length estimation

A visualization of the estimated localization of malignant tissue for an example biopsy is presented in Figure 2. The correlation between the cancer length estimates of the AI and the measurements of the pathologist was 0.96 (0.93 for positive cores). When aggregating the cancer extent of all cores within a case, the correlation was 0.98, both for all men and for men positive for cancer (Figure 3). An online tool (https://tissuumaps.research.it.uu.se/sthlm3/) allows interactive examination of predictions generated for 30 cores randomly selected (5 per ISUP score and 5 benign) from the test set.

## ISUP grading

The average pairwise kappa achieved by the AI relative to the panel of pathologists on the 87 Imagebase cases was 0.62. The pathologists had values ranging from 0.60 to 0.73, with the study pathologist (L.E.) having a kappa of 0.73. When considering a narrower grouping of ISUP grades (ISUP 1, ISUP 2-3 and ISUP 4-5), which often forms the basis for primary treatment selection, the AI scored even higher relative to the pathologists (Figure 4). The grades assigned by the panel and the AI to each Imagebase case are shown in Appendix (Figure S3).

The kappa obtained by the AI relative to the pathology report in the independent test set of 1,631 cores was 0.83 for all cores and 0.70 for positive cores only (see Appendix, Figure S4).

# Discussion

Grading prostate cancer can be a difficult procedure due to the complex nature of the score and its derivation. This has also been true for computer algorithms aiming at automating grading. The challenge is not only to develop an AI for this task, but also to demonstrate that it is consistent with current state-of-the-art diagnosis of prostate histopathology. Here, we have for the first time demonstrated AI-based grading of prostate biopsies on the level of leading urological pathologists represented by the ISUP Imagebase panel.

Due to the poor discriminative ability of the prostate specific antigen test and the systematic biopsy protocol of 10-12 needle cores, which is still in common usage, most biopsies encountered in clinical practice are of benign tissue. To reduce the workload of assessing these samples, we evaluated the AI's ability to assist the pathologist by pre-screening benign from malignant cores. With an estimated AUC of 0.997, the system could automatically remove 809 benign biopsies from 246 men without missing a single man out of the 211 with cancer diagnosed by the study pathologist (Figure 1). Since the pathology report was used as gold standard for this evaluation, the AI, by design, cannot achieve a higher sensitivity than the reporting pathologist. However, some malignant cores may still be overlooked by the pathologist but detected by the AI. As an illustration of this, Ozkan *et al.* evaluated the agreement of two pathologists in the assessment of cancer in biopsy cores.[5] Following examination of 407 cases, one pathologist found cancer in 231 cases, while the other found cancer in 202 cases. This suggests that an AI can not only streamline the workflow but could also improve sensitivity by detecting cancer foci that would otherwise be accidentally overlooked.

In this study, we have also evaluated the assessment of tumor burden (cancer length). We believe that both cancer detection and cancer length measurements can now be automated without sacrificing patient safety. In support of this and to provide interpretations of the DNN's predictions, we have published on our website high-resolution images of 30 cores randomly selected from the test data, accompanied by their ISUP grades and the AI's predictions.

The first attempt to use DNNs for the detection of cancer on prostate biopsies was reported by Litjens *et al.*[15] Using an approach similar to ours but based on a small dataset, they could safely exclude 32% of benign cores. A more recent study by Campanella *et al.* demonstrated an AUC of 0.977 for cancer detection.[16] There have also been attempts to undertake grading of prostate tissue derived from prostatectomy or based on tissue microarrays.[14,26] None of these studies achieved expert uro-pathologist level consistency in Gleason grading, estimated tumor burden or investigated the reproducibility of grading on needle biopsies, which are utilized for diagnostics in virtually every pathology laboratory worldwide. Moreover, no previous study has used a well-defined cohort of samples to estimate the clinical implications, with respect to key medical operating characteristic metrics such as sensitivity and specificity.[27]

The strengths of our study include the use of well-controlled, prospectively collected and population-based data covering a large random sample of men with both the urologists and the pathologist blinded to patient characteristics. Prostate cancers diagnosed in STHLM3 are representative for a screening-by-invitation setting, and the data include cancer variants that are notoriously difficult to diagnose (pseudohyperplastic and atrophic carcinoma), slides which required immunohistochemistry, and mimickers of cancer (Table S6). Despite these difficult cases, the AI achieved near perfect diagnostic concordance with the study pathologist. The study was subjected to a strict protocol, where the splitting of cases into training and test sets was performed at a patient level and all analyses were pre-specified prior to the evaluation of the independent test set, including code for producing tables, figures, and result statistics. A further strength is the use of Imagebase which is a unique dataset for testing the performance of the AI against highly experienced urological pathologists.

We trained the AI using annotations from a single, highly experienced urological pathologist (L.E.). The decision to rely on a single pathologist for model training was done to avoid presenting the AI with conflicting labels for the same morphological patterns and to thereby achieve more consistent predictions. L.E. has in several studies demonstrated high concordance with other experienced uro-pathologists, and therefore represents a good reference for model training.[28,29] For model evaluation, however, it is critical to assess performance against multiple pathologists (Figure 3).

The main limitation of this study is the lack of exact pixel-wise annotations, since the annotations may highlight regions that include a mixture of benign and malignant glands of different grades. To address this issue, we used a patch size large enough to cover glandular structures but small enough to minimize the presence of mixed grades within a patch, and we focused our attention on core and patient level performance metrics, which avoids caveats of patch-level evaluation and is clinically more meaningful.

# Conclusions

We have demonstrated that an AI based on DNNs can reach the level of highly experienced urological pathologists for the grading of prostate biopsies. We believe that the use of this system can increase sensitivity and promote patient safety by providing decision-support and by focusing the attention of the pathologist on regions of interest. In addition, the use of this system can reduce high intra-observer variability in the reporting of prostate histopathology by producing reproducible and consistent grading. A further benefit is that AI can provide diagnostic expertise in regions where this is currently unavailable. Our results warrant international validation, which we are pursuing in an ongoing project where we collect slides from seven countries.

# Author contributions

ME had full access to all the data in the study and takes responsibility for the integrity of the data and the accuracy of the data analysis. PS and KK contributed equally to algorithmic design and implementation, and drafting the manuscript. In addition, PS was mainly responsible for statistical analysis of results and KK was mainly responsible for high-performance computing. HO was mainly responsible for data management and participated in algorithmic design and implementation, and in drafting the manuscript. LS developed the online viewer application allowing visual examination of results. BD was involved in drafting the manuscript. BD, DMB, DGB, LE, AJE, DJG, PAH, KAI, JGK, GK, THVDK, KRML, JKMK, JO, CCP, HS, JRS, HT, TT, MV, MZ performed grading of the ImageBase dataset and provided pathology expertise and feedback. CL was involved in data collection. JL was involved in study design. PR and CW contributed to design and supervision of the study and to algorithmic design. In addition, PR contributed to high-performance computing and CW contributed to designing the online viewer. HG contributed to the conception, design and supervision of the study. MR contributed to the conception, design and supervision of the study and to algorithmic design. LE graded and annotated all the data used in the study, contributed to the conception, design, and supervision of the study, and helped draft the manuscript. ME was responsible for the conception, design and supervision of the study, and contributed to algorithmic design, analysis of results and drafting the manuscript. All authors participated in the critical revision and approval of the manuscript.

# Additional contributions

The Tampere Center for Scientific Computing and CSC - IT Center for Science, Finland are acknowledged for providing computational resources. The S:t Göran Hospital, Stockholm, is acknowledged for providing additional high-grade slides as training data. Carin Cavalli-Björkman, Britt-Marie Hune, Astrid Björklund, and Olof Cavalli-Björkman have been instrumental in logistical handling of the glass slides, and Masi Valkonen and Tomi Häkkinen in providing technical assistance. We thank the participants in the Stockholm-3 study for their participation.

# Funding

Funding was provided by the Swedish Research Council, Swedish Cancer Society, Swedish Research Council for Health, Working Life, and Welfare (FORTE), Academy of Finland [313921], Cancer Society of Finland, Emil Aaltonen Foundation, Finnish Foundation for Technology Promotion, Industrial Research Fund of Tampere University of Technology, KAUTE Foundation, Orion Research Foundation, Svenska Tekniska Vetenskapsakademien i Finland, Tampere University Foundation, Tampere University graduate school, The Finnish Society of Information Technology and Electronics, TUT on World Tour programme and the European Research Council (grant ERC-2015-CoG 682810). The funders had no role in study design, data collection, analysis and interpretation, writing of the report or making the decision to submit for publication.

|  | Stockholm-3 | Digitized (n=1,380) | | | |
| --- | --- | --- | --- | --- | --- |
|  | Biopsied (n=7,406) | Training (n=976) | Extra Training (n=93) | Test (n=246) | Imagebase (n=86) |
| Per Man | No. (%) | No. (%) | No. (%) | No. (%) | No. (%) |
| **Age†** | | | | | |
| <49 yr | 45 (0.61) | 4 (0.41) | 0 (0.0) | 1 (0.41) | 0 (0.0) |
| 50-54 yr | 639 (8.63) | 76 (7.81) | 2 (2.2) | 11 (4.47) | 10 (11.63) |
| 55-59 yr | 1221 (16.49) | 136 (13.98) | 4 (4.3) | 44 (17.89) | 8 (9.3) |
| 60-64 yr | 2027 (27.37) | 255 (26.21) | 8 (8.6) | 67 (27.24) | 23 (26.74) |
| 65-69 yr | 3294 (44.48) | 482 (49.54) | 17 (18.3) | 115 (46.75) | 44 (51.16) |
| ≥70 yr | 180 (2.43) | 20 (2.06) | 57 (61.3) | 8 (3.25) | 1 (1.16) |
| Missing | 0 (0.0) | 0 (0.0) | 5 (5.4) | 0 (0.0) | 0 (0.0) |
| **Previous negative biopsy** | | | | | |
| Yes | 505 (6.82) | 33 (3.39) | 0 (0.0) | 13 (5.28) | 7 (8.14) |
| No | 6901 (93.18) | 940 (96.61) | 0 (0.0) | 233 (94.72) | 79 (91.86) |
| Missing | 0 (0.0) | 0 (0.0) | 93 (100.0) | 0 (0.0) | 0 (0.0) |
| **PSA** | | | | | |
| <3 ng/mL | 1933 (26.1) | 228 (23.43) | 2 (2.2) | 43 (17.48) | 13 (15.12) |
| 3-5 ng/mL | 3458 (46.69) | 428 (43.99) | 2 (2.2) | 100 (40.65) | 48 (55.81) |
| 5-10 ng/mL | 1612 (21.77) | 213 (21.89) | 19 (20.4) | 73 (29.67) | 16 (18.6) |
| ≥10 ng/mL | 403 (5.44) | 104 (10.69) | 57 (61.3) | 30 (12.2) | 9 (10.47) |
| Missing | 0 (0.0) | 0 (0.0) | 13 (14.0) | 0 (0.0) | 0 (0.0) |
| **Digital rectal examination** | | | | | |
| Abnormal | 680 (9.18) | 133 (13.67) | 59 (72.0) | 39 (15.85) | 12 (13.95) |
| Normal | 6726 (90.82) | 840 (86.33) | 10 (12.2) | 207 (84.15) | 74 (86.05) |
| Missing | 0 (0.0) | 0 (0.0) | 13 (15.9) | 0 (0.0) | 0 (0.0) |
| **Prostate volume‡** | | | | | |
| <35 mL | 2701 (36.47) | 425 (43.68) | 26 (28.0) | 92 (37.4) | 42 (48.84) |
| 35-50 mL | 2494 (33.68) | 319 (32.79) | 18 (19.4) | 82 (33.33) | 36 (41.86) |
| ≥50 mL | 2211 (29.85) | 229 (23.54) | 22 (23.7) | 72 (29.27) | 8 (9.3) |
| Missing | 0 (0.0) | 0 (0.0) | 27 (29.0) | 0 (0.0) | 0 (0.0) |
| **Cancer length** | | | | | |
| No cancer | 4605 (62.18) | 139 (14.29) | 0 (0.0) | 35 (14.23) | 0 (0.0) |
| 0-1 mm | 545 (7.36) | 133 (13.67) | 2 (2.2) | 35 (14.23) | 4 (4.65) |
| 1-5 mm | 922 (12.45) | 258 (26.52) | 10 (10.8) | 61 (24.8) | 20 (23.26) |
| 5-10 mm | 449 (6.06) | 135 (13.87) | 17 (18.3) | 28 (11.38) | 20 (23.26) |
| >10 mm | 885 (11.95) | 308 (31.65) | 64 (68.8) | 87 (35.37) | 42 (48.84) |
| **Cancer grade** | | | | | |
| Benign | 4605 (62.18) | 139 (14.29) | 0 (0.0) | 35 (14.23) | |
| ISUP 1 (3 + 3) | 1558 (21.04) | 413 (42.45) | 1 (1.1) | 104 (42.28) | |
| ISUP 2 (3 + 4) | 761 (10.28) | 200 (20.55) | 1 (1.1) | 53 (21.54) | |
| ISUP 3 (4 + 3) | 253 (3.42) | 96 (9.87) | 1 (1.1) | 16 (6.5) | |
| ISUP 4 (4 + 4, 3 + 5 and 5 + 3) | 101 (1.36) | 63 (6.47) | 19 (20.4) | 21 (8.54) | |
| ISUP 5 (4 + 5, 5 + 4 and 5 + 5) | 128 (1.73) | 62 (6.37) | 71 (76.3) | 17 (6.91) | |
| | | | | | |
| Per Biopsy core | Biopsied (n=83,470) | Training (n=6,682) | Extra Training (n=271) | Test (n=1,631) | Imagebase (n=87) |
| **Cancer length** | | | | | |
| No cancer | 73595 (88.17) | 3724 (55.73) | 1 (0.37) | 910 (55.79) | 0 (0.0) |
| 0-1 mm | 3307 (3.96) | 915 (13.69) | 7 (2.58) | 203 (12.45) | 8 (9.2) |
| 1-5 mm | 4135 (4.95) | 1239 (18.54) | 41 (15.13) | 295 (18.09) | 44 (50.57) |
| 5-10 mm | 1822 (2.18) | 591 (8.84) | 85 (31.37) | 150 (9.2) | 24 (27.59) |
| >10 mm | 611 (0.73) | 213 (3.19) | 111 (40.96) | 73 (4.48) | 11 (12.64) |
| Missing | 0 (0.0) | 0 (0.0) | 26 (9.59) | 0 (0.0) | 0 (0.0) |
| **Cancer grade** | | | | | |
| Benign | 73595 (88.17) | 3724 (55.73) | 1 (0.37) | 910 (55.79) | |
| ISUP 1 (3 + 3) | 5664 (6.79) | 1530 (22.9) | 1 (0.37) | 349 (21.4) | |
| ISUP 2 (3 + 4) | 2051 (2.46) | 538 (8.05) | 1 (0.37) | 142 (8.71) | |
| ISUP 3 (4 + 3) | 903 (1.08) | 261 (3.91) | 2 (0.74) | 66 (4.05) | |
| ISUP 4 (4 + 4, 3 + 5 and 5 + 3) | 689 (0.83) | 424 (6.35) | 45 (16.61) | 92 (5.64) | |
| ISUP 5 (4 + 5, 5 + 4 and 5 + 5) | 568 (0.68) | 205 (3.07) | 221 (81.55) | 72 (4.41) | |

**Table 1:** Subject characteristics among all biopsied men in STHLM3 and among men whose biopsies were digitized, tabulated by men **(top)** and by individual biopsy cores **(bottom)**. No cancer grade information is shown for Imagebase, as the grading of this set of samples was performed independently by multiple observers. Imagebase cancer length was assessed by L.E.

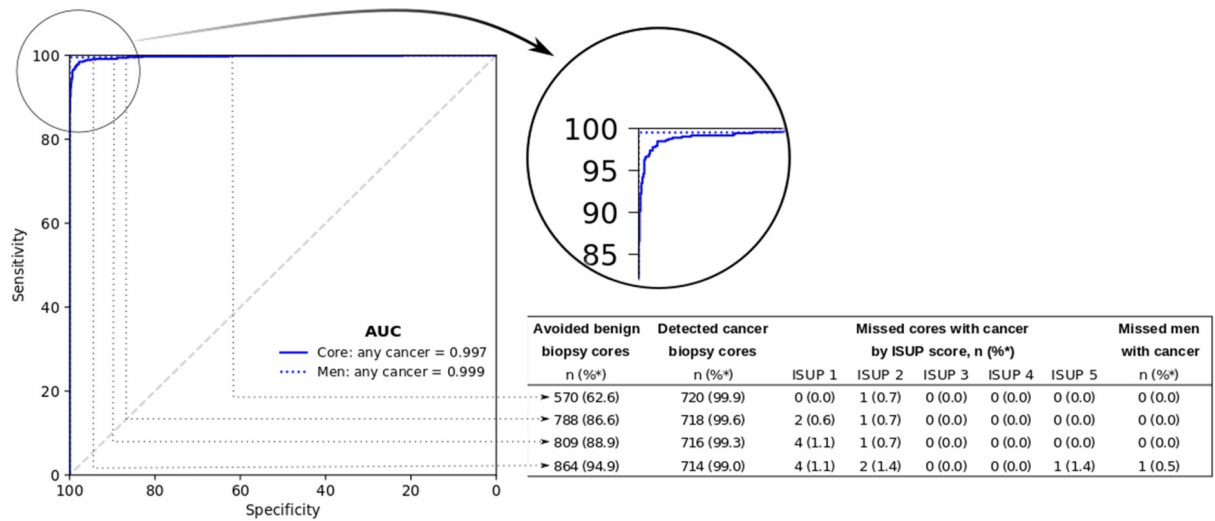

**Figure 1:** ROC and AUC for cancer detection **(left)**; by individual cores and by men. Four operating points on the core level curve are highlighted **(right)**. The first two columns from left show the number of biopsy cores that could be discarded from further consideration and the number of biopsy cores that would need pathological evaluation, respectively. The values in parentheses indicate the corresponding specificity and sensitivity. The next five columns show the number and percentage of missed malignant cores by ISUP score for each operating point. The rightmost column indicates the number and percentage of missed cancers among all men with cancer.

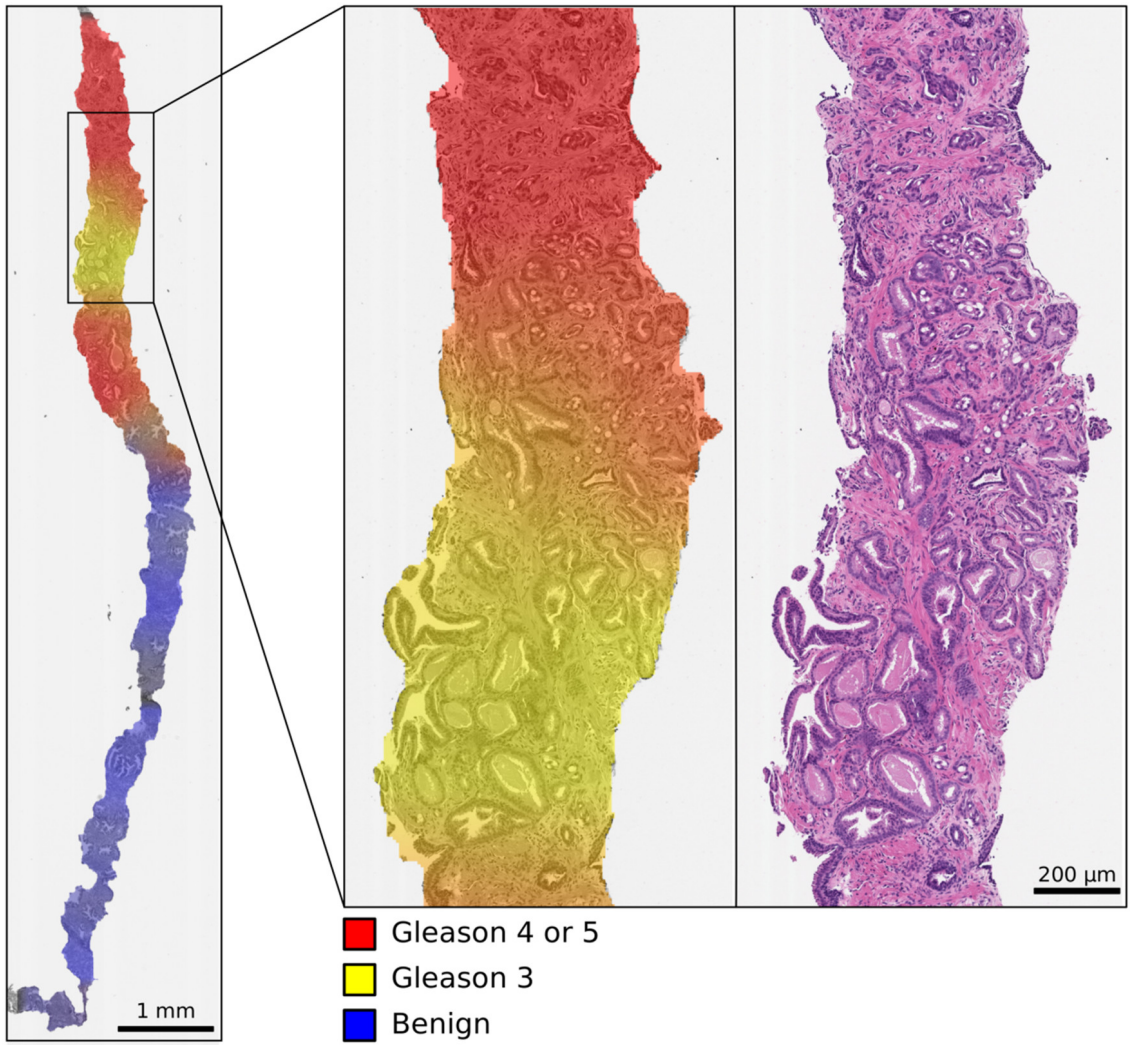

**Figure 2:** Color-coded visualization of cancer grades estimated by the AI. The colors represent the estimated probabilities for the presence of benign (blue), malignant low grade (Gleason 3, yellow) and malignant high grade (Gleason 4 or 5, red) tissue at different locations of the biopsy **(left)**. A magnified view of the AI output **(center)** and the corresponding H&E stained tissue **(right)** are shown for a region where an estimated transition between low- and high-grade morphology can be observed. This core from the test data was graded as ISUP 3 (GS 4+3) by the study pathologist.

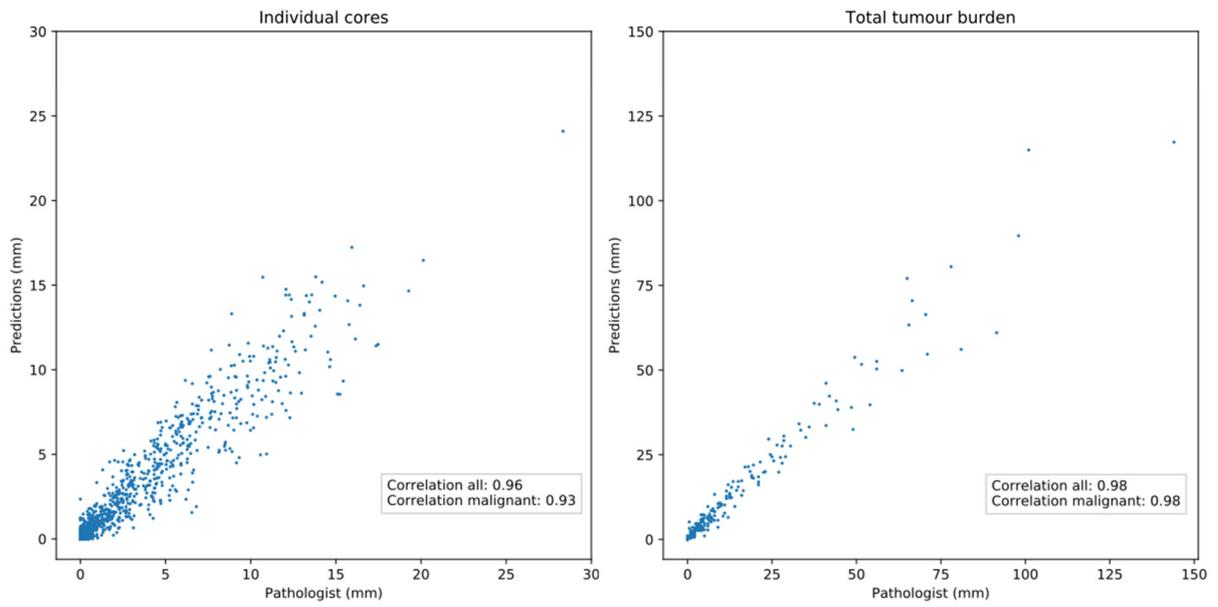

**Figure 3:** Scatterplots presenting the concordance between cancer lengths estimated by the AI and the pathologist for independent test data. Results are shown for individual cores **(left)** and aggregated over cores for each man **(right)**. Corresponding linear correlation coefficients computed for all cores and malignant cores only are shown in each plot. Data points in the left plot are jittered along the x-axis for clarity.

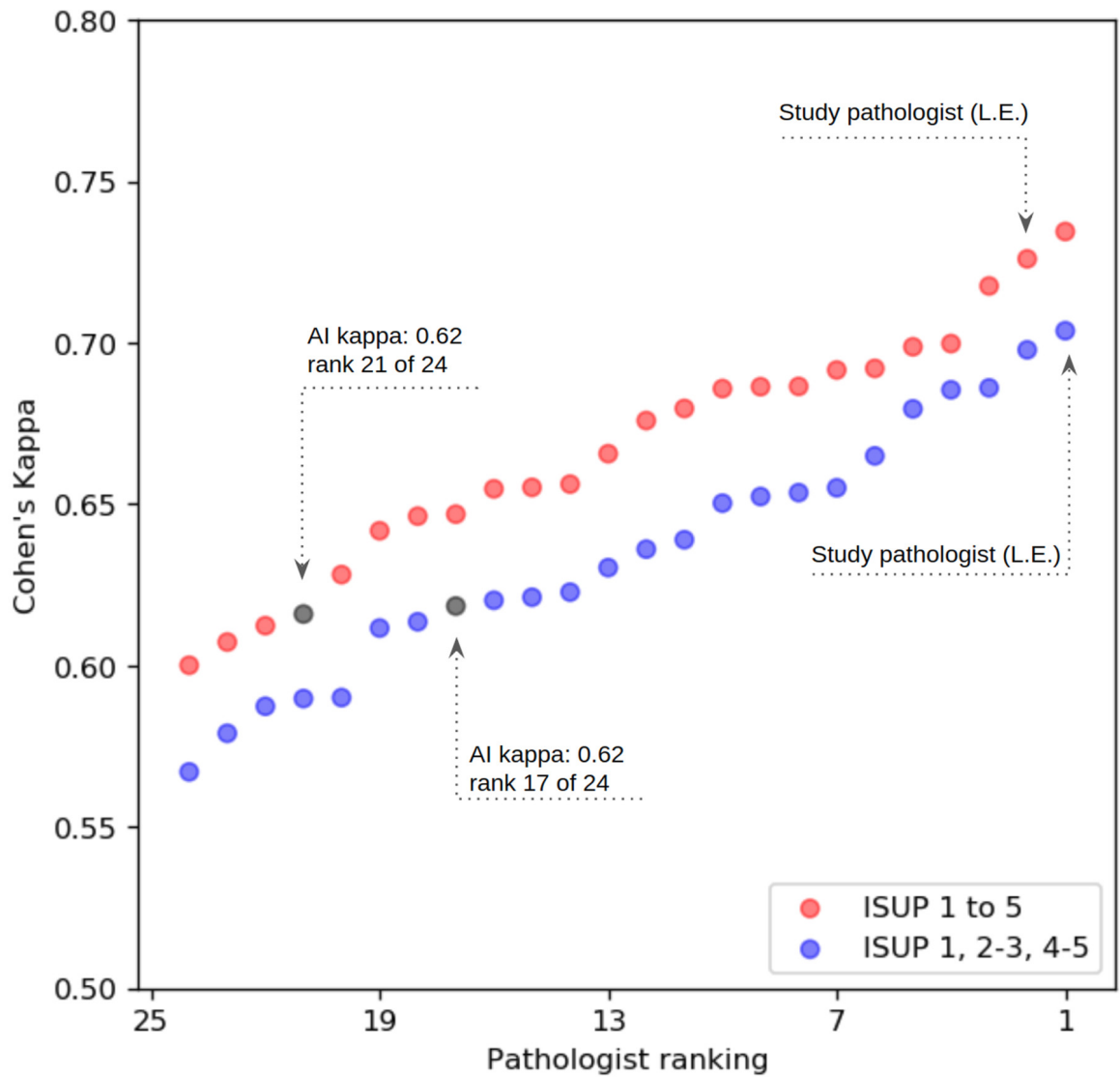

**Figure 4:** Cohen's kappa for each pathologist ranked from lowest to the highest. Each kappa value is the average pair-wise kappa for each of the pathologists compared against the others. To account for the natural order of the ISUP scores we used linear weights. The AI is highlighted with a black dot and an arrow. The study pathologist (L.E.) is highlighted with an arrow. Values computed based on all five ISUP scores are plotted in red, while values based on a grouping of ISUP scores commonly used for treatment decision are shown in blue.

# Appendix

## 1. Sample collection

Sample collection was carried out in two rounds. The first selection was the first 500 men with prostate cancer who were diagnosed in the Stockholm-3 study. All 10-12 cores from these men were scanned, in total 5,662 slides. The second round included all men with at least one core graded as Gleason Score (GS) 4+4 or 5+5 to enrich the amount of training examples from high grade biopsies. It also included a random selection of 497 men with at least one biopsy graded as GS 3+3. From each of these men, we included all positive cores and a randomly selected negative core. Finally, we randomly selected 139 cancer-free men from whom we included one randomly selected core. **Table 1** shows the distribution of grades among the selected men.

Due to the relatively few biopsy cores with GS 5+5 in the Stockholm-3 study, we collected an additional set of 271 slides from 93 men with high grade disease. Out of these slides, 23 were graded by Lars Egevad at Karolinska Hospital. The remaining 248 slides were collected in the Capio S:t Göran Hospital (Stockholm) between 2016-2017, and re-graded by Lars Egevad for the purpose of the AI development presented here. As these samples are not part of the Stockholm-3 cohort, we did not include them in the evaluation (i.e. in the independent test set), but only used them in training to improve the AI system's discrimination for high grade disease.

The linear cancer extent was generally measured from end to end in cases with discontinuous cancer. However, in cases with 1 or 2 cores infiltrated by low grade discontinuous cancer with a benign gap exceeding 3 mm, benign tissue was subtracted in the reporting of total cancer extent.

## 2. Image acquisition

The first round of slides was digitized using a Hamamatsu C9600-12 scanner and NDP.scan software v. 2.5.86 (Hamamatsu Photonics, Hamamatsu, Japan). The following batches of slides were scanned using an Aperio ScanScope AT2 scanner and Aperio Image Library v. 12.0.15 software (Leica Biosystems, Wetzlar, Germany). The pixel size at full-resolution (20X) was 0.45202 µm (Hamamatsu) or 0.5032 µm (Aperio). The resulting RGB images were stored at 8 bits per channel in NDPI (Hamamatsu) or SVS (Aperio) format.

## 3. Hardware and software

Computations were performed on two graphics processing unit (GPU) clusters (Tampere Center for Scientific Computing, Finland and CSC IT Center for Science, Finland), utilizing a total of 136 x Tesla P100 GPUs (Nvidia, Santa Clara, CA, USA), distributed on 37 nodes. The GPUs were running Nvidia driver v. 410.79, CUDA v. 9.2.148 and cuDNN v. 7.3.1. The nodes were equipped with either a 20-core Xeon E5-2640 v4 or a 24-core Xeon E5-2680 v4 CPU (Intel, Santa Clara, CA, USA), 256 or 512 GB RAM and 1.2 or 1.6 TB of local SSD storage. We used 32-bit floating point precision for GPU computation.

Images were accessed using OpenSlide[1] (v. 3.4.1) via the Python interface (v. 1.1.1) and an interface[2] for MATLAB (The MathWorks, Natick, MA, USA). Pre-processing steps for image segmentation and label digitization were performed using MATLAB R2017b. Deep neural networks (DNN) were implemented in Python 3.6.4 using Keras[3] (v. 2.2.4) with the TensorFlow[4] (v. 1.11.0) backend, and all boosted trees using the Python interface for XGBoost (v. 0.72).

## 4. Image pre-processing

### 4.1. Segmentation of tissue

Our image pre-processing workflow is depicted in **Figure S1**. First, we employed a Laplacian filtering algorithm to separate tissue from background and pen mark annotations. We first read images downsampled by a factor of 16 directly from the resolution pyramids present in the image files using Openslide[1]. We then converted the images from RGB to grayscale in accordance with the NTSC standard by calculating the weighted sum 0.2989 x R + 0.5870 x G + 0.1140 x B for each pixel. A filter approximating the 2D Laplacian operator, implemented in MATLAB's *fspecial* function, was applied on the image and the absolute magnitude of the resulting response was thresholded using Otsu's method[5]. This results in a binary mask indicating the locations of regions with variation in pixel intensity, mainly corresponding to in-focus tissue. The regions were smoothed by applying morphological closing with a disk-shaped structuring element having a radius of 50 µm, followed by filling of holes and removal of objects having an area smaller than 100 000 µm$^2$. Any remaining pen marks were removed based on their color by performing the HSV transform and excluding any objects whose mean hue was less than 0.7.

The final binary tissue masks *T* were rescaled back to full resolution using nearest neighbor interpolation, followed by lossless PackBits compression, and storage in Tiff or BigTiff format (depending on image dimensions) using a tile size of 1024 and 'Chunky' planar configuration.

## 4.2. Segmentation of pen marks

To segment the penmark annotations, downsampled images were first read and converted to grayscale as described above for the tissue segmentation. Dark regions were extracted by applying Otsu's thresholding[5], followed by taking the complement of the resulting binary mask and applying morphological closing with a disk-shaped structuring element having a radius of 50 µm. Candidate pen mark regions were obtained via a pixel-wise logical operation as the set of pixels belonging to the dark regions but not belonging to the tissue regions. The regions were further refined by filling of holes and removal of objects having an area smaller than 100 000 µm$^2$ or width less than 400 µm. The minimal width criterion is needed to remove the long but thin slide edges visible in some WSIs. Any remaining out-of-focus tissue regions falsely detected as pen marks were removed based on their color by performing the HSV transform and excluding any objects, whose mean hue exceeded 0.6.

The final binary pen masks *P* were rescaled back to full resolution using nearest neighbor interpolation, followed by lossless PackBits compression, and storage in Tiff or BigTiff format (depending on image dimensions) using a tile size of 1024 and 'Chunky' planar configuration.

## 4.3. Digitization of annotations

Relying on the tissue masks *T* and pen masks *P*, we generated digital pixel-wise label masks *L*, indicating whether each pixel represents background, normal tissue or cancer tissue. For this purpose, the tissue and pen masks were utilized at a resolution corresponding to downsampling by a factor of 16 relative to full resolution. A smoothed version of the tissue mask $T_s$ was generated by applying morphological closing with a disk-shaped structuring element having a radius of 100 µm, followed by filling of holes.

In order to populate a binary mask **C** indicating tissue pixels labeled as cancerous, each pen mark region $P_i$ in pen mask **P** was processed according to the following algorithm:

1. Smooth $P_i$ by morphological closing with a disk-shaped structuring element (radius 100 μm) and fill holes.
2. Set any overlapping tissue and pen mark pixels $T_s \cap P_i$ to FALSE in both masks.
3. Find the set of pixels at tissue boundaries $T_B$.
4. Find the set of pixels at the boundaries of this pen mark $P_B$.
5. For each pixel in $P_B$, find nearest pixel in $T_B$ via Kd-tree search[6] to construct the set of nearest tissue pixels $T_N$.
6. Discard all pixel pairs in $P_B$ and $T_N$ located further than 2000 μm from each other.
7. Compute the mean direction $\theta$ and circular variance $\sigma$ of the vectors defined by starting points in $P_B$ and end points in $T_N$[7].
8. Discard all pixel pairs $j$ in $P_B$ and $T_N$ where the direction of the vector $T_{N,j} - P_{B,j}$ differs from $\theta$ by more than $\theta_t$, where $\theta_t$ is an angle of 20° or $\pi\sigma$ radians, whichever is larger. If no pairs satisfy the criterion, keep the two pairs with direction closest to $\theta$.
9. For each pixel $j$ in $T_N$, attempt to locate the pixel on the opposite side of the tissue section by starting from $T_{N,j}$ and proceeding in the direction defined by the vector $T_{N,j} - P_{B,j}$ until the first non-tissue pixel is encountered or the distance from $P_{B,j}$ exceeds 2000 μm. Collect the located pixels into the set $T_O$.
10. Get the convex hull of the set of points $T_N \cup T_O$ as a binary image **H** and thicken it by 3 pixels to rectify interpolation errors.
11. Assign the set of pixels $H \cap T$ a value of TRUE in **C**.

Based on the mask **C**, annotated tissue pixels were assigned the label 2 (i.e. cancer) in the label mask **L**. Non-annotated tissue pixels located on the same tissue sections as annotated tissue were assigned the label 1 (i.e. normal). Non-tissue regions and tissue regions not containing any annotated tissue were assigned the label 0 (i.e. background/unknown). The slides in our dataset typically contain two tissue sections, only one of which is annotated with pen marks by a pathologist, and we therefore chose to disregard the unannotated section to reduce label uncertainty.

To avoid regions with locally inconsistent labels, we refined the initial label mask **L** by first applying morphological dilation with a disk-shaped structuring element (radius 100 μm) to the regions labeled as cancer, and then applied the same operation to the regions labeled as normal tissue. Finally, we introduced some margin at the boundaries of cancerous tissue by dilating the regions labeled as cancer with a disk-shaped structuring element (radius 700 µm) and assigning the label 0 (i.e. unknown) to all tissue pixels within the expanded region.

For slides containing only benign tissue and hence no pen marks, we directly assigned the label 1 to all tissue pixels indicated by *T* instead of applying the algorithm described above.

For a subset of 62 slides with grade-specific annotations relying on color-coded pen marks, we used a modified version of the above algorithm. For each pen mark region $P_i$ on these slides, we computed the mean RGB value of the region from the original image. We then computed the Euclidean distance between this RGB vector and each of the possible colors: green (0, 255, 0) for Gleason 3, blue (0, 0, 255) for Gleason 4 and black (0, 0, 0) for Gleason 5. The color producing the shortest distance was used as the basis of labeling the corresponding tissue region. That is, instead of assigning annotated tissue pixels the label 2 (i.e. cancer) in the label mask *L*, we assigned the values 3 (Gleason 3), 4 (Gleason 4) or 5 (Gleason 5). Pixels with conflicting labels indicated by multiple, differently colored pen marks were assigned the label 255 (mixed Gleason).

The final unsigned 8-bit integer label masks *L* were rescaled back to full resolution using nearest neighbor interpolation, followed by lossless LZW compression, and storage in Tiff or BigTiff format (depending on image dimensions) using a tile size of 1024 and 'Chunky' planar configuration.

## 4.4. Extracting patches from the WSI

To extract image regions, which are small enough to mainly represent only one class of tissue and to fit in GPU memory, we divided each WSI into smaller, partially overlapping patches. Several preprocessing operations were simultaneously performed: exclusion of non-tissue regions, assignment of class labels for each patch, and adjustment of pixel size. We used the pre-generated tissue masks *T*, pen masks *P* and label masks *L* as the basis of this process. The masks were downsampled by a factor of 16 via nearest-neighbor interpolation to speed up processing while retaining their original level of fidelity. Patch extraction was run in parallel on multiple CPU cores.

Each WSI was processed following a sliding-window approach with tunable stride *s*, patch size *t* and resolution level *r*. The resolution level corresponds to a downsampling factor of $2^r$ relative to full resolution and an optical magnification of $20/2^r$. We used *s* = 299 px (approx. 135 µm), *r* = 1 (pixel size approx. 0.90 µm) and experimented with different values for *t* ranging from 598 px (approx. 270 μm) to 1196 px (approx. 540 μm). For each window, we first extracted the corresponding region from *T*. If less than 50% of the pixels within the region were indicated to be tissue, we skipped to the next window. Otherwise, we extracted the same region from *L* and assigned the patch a label according to the mode of tissue pixel labels in the label mask. We observed improving classification performance with increasing patch size, up until 1196 px (540 μm), which was the largest size feasible in terms of available GPU memory (data not shown).

We then read the image data of the patch at the specified location and resolution level directly from the resolution pyramids present in the image files using Openslide[1]. If the desired level *r* was not present in the image file, we read the image at the next available higher resolution level. The patch was then resized to its final dimensions, at the specified resolution level, via Lanczos interpolation. At this point we also compensated for the different pixel size of the Hamamatsu and Aperio scanners by upsampling patches from Aperio slides to match the higher resolution of Hamamatsu. As a result, all produced patches had the same pixel size and pixel dimensions, each covering a physical region of identical size in micrometers. Any pixels in the patch indicated as non-tissue in *T* or as pen marks in *P* were then assigned a constant value corresponding to fully white (255, 255, 255) to remove any background patterns and annotation information from the image data. Finally, the patches were compressed using JPEG compression (quality 80) and stored on disk, either as .jpg files (one file per patch) or as .tar archives (one file per slide), depending on disk hardware used in downstream steps.

## 5. Data management and quality control

Prior to image preprocessing, all WSIs were visually examined to exclude slides unsuitable for analysis. The excluded slides included 86 slides representing immunohistochemical instead of H&E staining, 8 slides with failed H&E staining resulting in near complete lack of stain, 3 slides with corrupted data, and 23 slides containing smeared pen marks spread on top of tissue regions.

All tissue, pen and label masks were visually examined to verify that the segmentation and digital annotations were accurate. Masks with inaccuracies were refined manually using a purpose-built interactive tool in MATLAB. For a total of 10,016 processed slides, 81 tissue masks (0.8%), 109 pen masks (1.1%) and 142 label masks (1.4%) required manual corrections. The corrections were mainly due to debris falsely detected as tissue, faint pen marks missed by the algorithm, and slide labels or other text located close to slide edges falsely detected as pen mark annotations. The presence or absence of pen marks, indicative of cancerous or benign slides, respectively, was confirmed visually and checked for consistency with the clinical data.

A total of 10,185 cores were digitized during multiple rounds of scanning **(Table S1)**. We developed a database for organizing the images together with clinical information. The filenames of the scanned slides contain the study specific identification variable that is used to link the images to the clinical data. A total of 268 (2.6%) images from scanning had corrupted filenames, which made linkage to clinical data impossible. There was an overlap of slides that were rescanned between rounds of scanning, and in these cases we kept the slides from the first scanning round and excluded 613 rescanned slides. Furthermore, we excluded 45 slides that were duplicated within the same scanning round and 298 slides where there was an inconsistency between the cancer pen mark on the image and the cancer indicator variable in the clinical data.

# 6. Patch-level classifier

## 6.1. Model

We used a two-stage model for classifying individual image patches (see **Figure S2**). The first stage of the model classifies image patches in binary fashion as either benign or cancerous, while the second stage performs Gleason grading. We included the benign class also into the second stage model in order to obtain a richer representation of the data, and to allow optionally applying or evaluating the two model stages separately of each other. The utility of the two-stage approach is three-fold: 1) it allowed us to fully utilize all training data, 2) it allowed uncoupling the training of models for the detection and grading tasks, which require different numbers of training epochs to avoid overfitting, and 3) it enables adjusting the classifier's operating point for the cancer detection task in a straightforward manner, independently of the Gleason grading task (see Section 6.2. for details).

We evaluated the following convolutional neural network (CNN) architectures: Inception V3[8], ResNet50[9] Inception-ResNet V2[10] and Xception[11]. Based on evaluation on validation data (a temporary split of the training data to enable independent evaluation of different models' performance without testing on the independent test set), Inception V3 offered the best performance in terms of classification accuracy (see Section 8.1; **Table S2**). In all cases, we applied global average pooling to the output of the last convolutional layer, followed by a dense layer with softmax activation to output a vector of estimated probabilities over all classes. The input shape was matched to the patch size (typically 598 x 598 pixels).

To improve generalization, we used an ensemble of multiple CNN models at each of the two stages. All these CNNs share the same architecture, and variability within the ensembles arises from the stochastic sampling during training. We used ensembles of 5 CNNs for cross validation experiments on training data and observed improved performance compared to using a single CNN (see Section 8.2.; **Table S3**). When training the final ensembles on all training data, we used 30 CNNs for each of the two stages.

## 6.2. Training

For training the binary first stage models, we used all benign and cancer patches from all training slides, irrespective of Gleason grade. For training the second stage models, we used a four class setting: benign, and Gleason patterns 3, 4 and 5. We used benign patches from all training slides, and inferred the Gleason grade label of each cancerous patch from the corresponding clinical data. Due to the lack of annotations indicating where each Gleason pattern is located on slides with multiple patterns (e.g. 3+4), we only used cancerous patches from slides with a single Gleason pattern (e.g. 3+3) for training. As an exception to this rule, we assigned patch-level Gleason labels directly based on grade-specific annotations where available (62 slides). In these cases, the maximum of the slide-level Gleason grades was assigned to patches with a mixed Gleason label. At both stages, we excluded patches with the label 'unknown' from training.

As an alternative to discarding slides with multiple Gleason patterns when training the second stage model, we evaluated an iterative approach. This involved: 1) training a model only on slides featuring a single Gleason pattern, 2) applying this model to predict the patch-level Gleason grades for slides featuring multiple patterns and 3) either fine-tuning the model further or training a new model from scratch using the predicted grades as additional training labels. Moreover, we experimented with using the primary Gleason grade (e.g. 3 for a 3+4 slide) as the label for all patches from a slide, or randomly choosing the Gleason grade for each patch from the two slide-level grades. Our experiments indicated that none of these approaches improved classification performance compared to only using single Gleason slides as training data (data not shown).

In order to compensate for the considerably imbalanced distribution of different classes in the training data, which can be detrimental for CNN models[12], we performed class balancing via subsampling before each 'epoch'. Specifically, we always included all training patches representing the rarest class, and randomly sampled the same amount of patches from all other classes. This results in a uniform class distribution without duplicated examples. The model was then trained until this set of patches was exhausted, and the random sampling step was repeated with replacement for the next 'epoch'. In our terminology, an epoch thus refers to a full pass through the minority class training examples. We trained the first stage models for 15 epochs and the second stage models for 60 epochs (see Section 8.3.; **Table S4-S5**).

To improve generalization and to obtain rotational invariance, each patch was randomly rotated by either 0°, 90°, 180° or 270°, and then flipped vertically with a probability of 50% before being fed to the CNN. This data augmentation step was repeated every time a patch was drawn. As a result of this approach, each patch was presented to the CNN in different, randomly sampled orientations at different epochs, without increasing the overall number of patches processed during a single epoch.

For optimization, we employed the Adam[13] algorithm using the following default parameters: learning rate = 0.001, $\beta_1$ = 0.9, $\beta_2$ = 0.999, $\varepsilon$ = $10^{-7}$, decay = 0 and categorical cross entropy as the loss function. We evaluated initialization with random weights or weights trained on ImageNet[14], and opted for the latter due to faster convergence. We used a minibatch size of 32 for Inception V3, and 16 for all other architectures, as dictated by the amount of available GPU memory.

During training, we employed multiple GPUs following the data parallelism strategy implemented in Keras. More specifically, the model was replicated on each GPU, each minibatch was divided into multiple sub-batches, and each one of them was processed by one GPU. The results were then concatenated on the CPU and the CNN weights were synchronously updated for all model replicas. We typically used two parallel GPUs per training run. Model training took approximately 5 or 3 days per one first stage or second stage CNN, respectively.

## 6.3. Prediction

In the prediction phase, we applied each trained model to all patches from slides representing the test set. No test-time data augmentation was used. This resulted in a set of predicted class-wise probabilities output by each model for each patch. These probabilities were further used as input for slide-level classification (see Section 7).

## 6.4. Visualization

For the purpose of visualization in the online viewer (accessible at *https://tissuumaps.research.it.uu.se/sthlm3*), we first computed the mean of the predicted probabilities over all the models in the ensemble for each patch, resulting in a single vector of class-wise probabilities per patch. Due to partial overlap between the patches, each pixel in the WSI is covered by multiple patches. Therefore, for each pixel, we computed the mean probabilities over all patches associated with the pixel, resulting in a single vector of class-wise probabilities per pixel. Bilinear interpolation was used to scale up the array of probabilities to match the dimensions of the full-resolution WSI.

To represent and store the data as an RGB image, hereafter referred to as a confidence mask, we constructed the channels as follows:
1. Cancer of any grade: The complement of the probability of the benign class predicted by the first stage ensemble.
2. Low grade cancer: The probability of Gleason grade 3 predicted by the second stage ensemble.
3. High grade cancer: The sum of the probabilities of Gleason grade 4 and 5 predicted by the second stage ensemble.

The values were converted from floating point to uint8 precision and the confidence mask was stored in TIFF format using LZW compression. The confidence mask was then tiled into pyramidal DZI format using vips[15] with a tile size of 512, overlap of 16 and PNG compression. The original WSI was tiled similarly to the confidence mask, except for using JPEG compression (quality 80).

# 7. Slide-level classifier

## 7.1. Model

We employed a model-based approach relying on boosted trees, implemented using XGBoost[16], for aggregating patch-level predictions into slide-level predictions (see **Figure S2**). We trained one boosted tree classifier based on the patch-level predictions of each CNN, thus forming ensembles of boosted trees. We trained one ensemble for each slide-level prediction task: 1) cancer detection (based on the first-stage CNN ensemble) and 2) estimation of cancer length (based on the first-stage CNN ensemble), and 3) Gleason grading (based on the second stage CNN ensemble).

For a slide with $n$ patches, we extracted slide-level features from the $n \times c$ matrix of class-wise probabilities estimated by a CNN for $c$ classes ($c = 2$ for the first stage CNNs, $c = 4$ for the second stage CNNs).

The following features were computed for each slide in class-wise manner over all $n$ patches, each resulting in a $1 \times c$ vector:
- Sum
- Median
- Maximum
- 99.75th percentile
- 99.50th percentile
- 99.25th percentile
- 99th percentile
- 98th percentile
- 95th percentile
- 90th percentile
- 80th percentile
- 10th percentile
- Number of patches with probability > 0.999
- Number of patches with probability > 0.99
- Number of patches with probability > 0.9

The following features were additionally computed for each slide:
- Total number of patches $n$ (1x1)
- Number of patches where each class had the highest probability ($1 \times c$)

## 7.2. Training

For training ensembles of boosted trees for cancer detection and cancer length estimation, we used features (see Section 7.1.) computed from the predictions generated for all training slides by the 30 CNNs in the cancer detection ensemble. For cancer detection, we used the binary logistic ('binary:logistic') loss function, a maximum tree depth of 5, and 100 iterations.

For the cancer length estimation, we used the linear regression ('reg:linear') loss function, a maximum tree depth of 2, and 200 iterations.

For training an ensemble of boosted trees for ISUP grading, we used features extracted from the predictions generated for all training slides by the 30 CNNs in the grading ensemble. We used the softmax ('multi:softprob') loss function, a maximum tree depth of 3, and 300 iterations. To handle class imbalance, we applied sample weights, which were inversely proportional to the number of training samples representing each ISUP class.

In order to select the maximum tree depth and the number of trees, which we considered the key parameters for the boosted tree models, we performed a parameter grid search. The grid search was evaluated on a 5-fold patient-level cross validation on the training data, with an ensemble of 5 cancer detection CNNs and 5 grading CNNs trained for each fold (see Section 8.4.; **Figure S5-S7**). For other parameters, we used the following (default) values: booster: gbtree; eta: 0.3; gamma: 0; min_child_weight: 1; max_delta_step: 0; subsample: 1; colsample_bytree: 1; colsample_bylevel: 1; lambda: 1; alpha: 0; tree_method: auto; sketch_eps: 0.03; scale_pos_weight: 1; updater: [grow_colmaker, prune]; refresh_leaf: 1; process_type: default; grow_policy: depthwise; max_leaves: 0; max_bin: 256; predictor: cpu_predictor.

## 7.3. Prediction

In the slide-level prediction phase, the patch-level predictions generated by the CNN ensembles (see Section 6.3.) were aggregated into slide-level features (see Section 7.1) and provided as input to the trained ensembles of boosted trees (see Section 7.2).

Probability of cancer of any grade being present was computed for each slide as the mean of the probabilities output by the 30 boosted trees comprising the cancer detection ensemble. A final estimate of cancer length for each slide was obtained by computing the mean of the predictions generated by the 30 boosted trees in the length estimation ensemble. Any negative values were replaced by zero. Grade-specific probabilities of cancer were computed for each slide as the mean of the probabilities output by the 30 boosted trees comprising the grading ensemble.

We obtained a final classification outcome for each slide by assigning the slide as benign if the average probability estimated by the cancer detection ensemble was below a chosen threshold corresponding to an operating point of choice. The grading results presented in this study are based on an operating point corresponding to a sensitivity of 99%. For slides with probabilities exceeding the threshold and thus classified as malignant, the ISUP score was assigned based on a Bayesian decision rule. The aim of using this approach instead of a simple argmax-rule, based on the estimated probabilities alone, was to encourage assigning higher rather than lower ISUP scores and to recognize the varying severity of misclassifications on the ordinal ISUP scale. More specifically, we assigned the grade to each slide predicted as malignant such that:

$$argmin_a \ R(a \mid x)$$

that is, given the input class-wise probabilities *x*, we choose the grade *a* that minimizes the conditional risk:

$$R(a \mid x) = \sum_y p(y \mid x) \, L(y, a)$$

where *L(y,a)* defines the loss associated with assigning grade *a* when the true grade is *y*. We defined our loss matrix *L* such that:

$$L(y, a) = \begin{cases} 0.1 \, |y - a|, & a > y \\ 0.2 \, |y - a|, & a < y \end{cases}$$

that is, underestimating a grade is perceived as twice as costly as overestimating a grade, and the cost increases linearly with the magnitude of the error. It should be noted, that this choice of preferring overestimation of ISUP score over underestimation by a factor of two is arbitrary. We based our decision on considerations of the clinical consequences of underestimation relative to overestimation, and on cross-validation experiments.

# 8. Supplementary results

## 8.1. Model architecture comparison

We compared different CNN architectures in terms of their performance based on a single validation split, where a random selection of 20% and 80% of the men in training data were allocated for validation and training, respectively (i.e. no data from the test set was used for these experiments). In contrast to the data used for subsequent cross validation experiments and final model training, the dataset used in this experiment did not yet include the additional set of 271 slides from 93 men with high grade disease collected outside the Stockholm-3 trial. Therefore, for the purposes of this experiment, we pooled together all patches representing an ISUP score of 5. That is, malignant patches from slides with a Gleason score of 4+5, 5+4 or 5+5 were all used as examples of Gleason pattern 5 in training.

We included the following architectures in the comparison: Inception V3[8], ResNet50[9], Inception-ResNet V2[10] and Xception[11]. For each architecture, we trained an ensemble of 5 CNNs for various numbers of epochs (see Section 6 for details). In this experiment, we used a single CNN ensemble for cancer detection, cancer length estimation and grading. For each architecture and number of epochs, we then trained three ensembles of boosted trees for cancer detection, cancer length estimation and grading, using the patch-level predictions from training slides as input (see Section 7 for details).

We evaluated the performance of each CNN architecture in: 1) cancer detection based on the AUC of discriminating malignant from benign biopsies, 2) cancer length estimation based on the linear correlation coefficient between estimated and reported lengths, and 3) grading based on Cohen's kappa computed between the ISUP grades assigned by the AI and the study pathologist for biopsies indicated as malignant by the pathologist (**Table S2**). The results indicated that out of the evaluated architectures, Inception V3 achieved the best performance in all three tasks.

## 8.2. Model ensembling

We evaluated the value of ensembling multiple CNNs to improve generalization based on a 5-fold patient-level cross validation on training data. For each fold, we used patches from the slides of the training patients in that fold to train an ensemble of 5 Inception V3 models for cancer detection and length estimation (see Section 6 for details). We trained all models for 15 epochs. For each fold, we then trained two ensembles of boosted trees for cancer detection and cancer length estimation, using the patch-level predictions generated by 1-5 CNNs for training slides as input (see Section 7 for details).

The trained patch- and slide-level ensembles were then applied to generate predictions for the validation slides in each fold, and the results were aggregated over all five folds. We evaluated performance as a function of ensemble size in cancer detection based on the AUC of discriminating malignant from benign biopsies, and in cancer length estimation based on the linear correlation coefficient between estimated and reported lengths (see **Table S3**). We observed a trend of increasing AUC and correlation with increasing ensemble size.

## 8.3. Number of epochs

We evaluated the optimal number of training epochs for the first stage ensemble of cancer detection CNNs and the second stage ensemble of cancer grading CNNs, based on a 5-fold patient-level cross validation on training data. For each fold, we used patches from the slides of the training patients in that fold to train two ensembles, each consisting of 5 Inception V3 models (see Section 6 for details). We then trained three ensembles of boosted trees for cancer detection, cancer length estimation and grading, using the patch-level predictions from training slides as input (see Section 7 for details). The trained patch- and slide-level ensembles were then applied to generate predictions for the validation slides in each fold, and the results were aggregated over all five folds. The entire process was repeated using either 10, 15 or 20 epochs for training the first stage ensemble, and either 40, 60 or 80 epochs for training the second stage ensemble.

One should note that the amount and class distribution of available training data is different for the two stages, which means that the epoch numbers are not comparable between the two ensembles. It is also worth pointing out, that our sampling over the number of epochs is very sparse compared to usual CNN training curves, where training and validation accuracy are typically presented for every epoch. The reason for this is that evaluating the performance of our system at the slide level requires: 1) applying all trained CNNs comprising the two ensembles to generate predictions for all tissue patches in the training slides, 2) training an ensemble of boosted trees, and 3) applying both the CNN ensembles and the boosted tree ensembles to generate predictions for validation slides. The computational cost of repeating this process after every epoch in a cross-validation setting would have been prohibitively high. An alternative would have been to assess training curves based on patch-level metrics computed on a limited set of validation patches. However, based on our observations, drawing conclusions on slide level performance based only on patch-level metrics is not straightforward, mainly due to the considerable amount of noise associated with our pixel-level and patch-level ground truth annotations. We therefore chose to evaluate the performance of the system with a sparse sampling over epochs, but with a focus on the slide-level metrics, which are easy to interpret and relevant in view of the system's potential clinical utility.

We evaluated the effect of the number of epochs on: 1) cancer detection based on the AUC of discriminating malignant from benign biopsies (**Table S4**), 2) cancer length estimation based on the linear correlation coefficient between estimated and reported lengths (**Table S4**), and 3) grading based on Cohen's kappa computed between the ISUP grades assigned by the AI and the study pathologist for biopsies indicated as malignant by the pathologist (**Table S5**). Based on these results, we selected 15 and 60 as the optimal number of epochs for the first and the second stage CNN ensembles, respectively.

## 8.4. Boosted tree hyperparameter selection

To select hyperparameter values for boosted trees, we performed a grid search over different tree numbers and tree depths (see Section 7) and evaluated them on a 5-fold patient-level cross validation on training data. More specifically, we used the patches from slides corresponding to the training patients of each fold to train an ensemble of 5 Inception V3 models for cancer detection and cancer length estimation, and an ensemble of 5 Inception V3 models for grading (see Section 6 for details). That is, we trained a total of 50 CNNs for this experiment. The cancer detection models were trained for 15 epochs and the grading models for 60 epochs.

For each fold, we then trained three ensembles of boosted trees for cancer detection, cancer length estimation and grading, using the patch-level predictions from training slides as input (see Section 7 for details). The training of these boosted trees was repeated with different combinations of tree depths (2-5), and number of trees (20-400). The trained patch- and slide-level ensembles were then applied to generate predictions for the validation slides in each fold, and the results were aggregated over all five folds. We thus obtained a set of predictions for all slides in our training data, based on different tree depth and tree number values.

We evaluated the effects of these parameter choices on: 1) cancer detection based on the AUC of discriminating malignant from benign biopsies (**Figure S5**), 2) cancer length estimation based on the linear correlation coefficient between estimated and reported lengths (**Figure S6**), and 3) grading based on Cohen's kappa computed between the ISUP grades assigned by the AI and the study pathologist for biopsies indicated as malignant by the pathologist (**Figure S7**). For each of the three tasks, we chose parameter values corresponding to relatively stable regions of high AUC, correlation and Cohen's kappa, respectively, to be used for the final model training.

# 9. Supplementary Figures

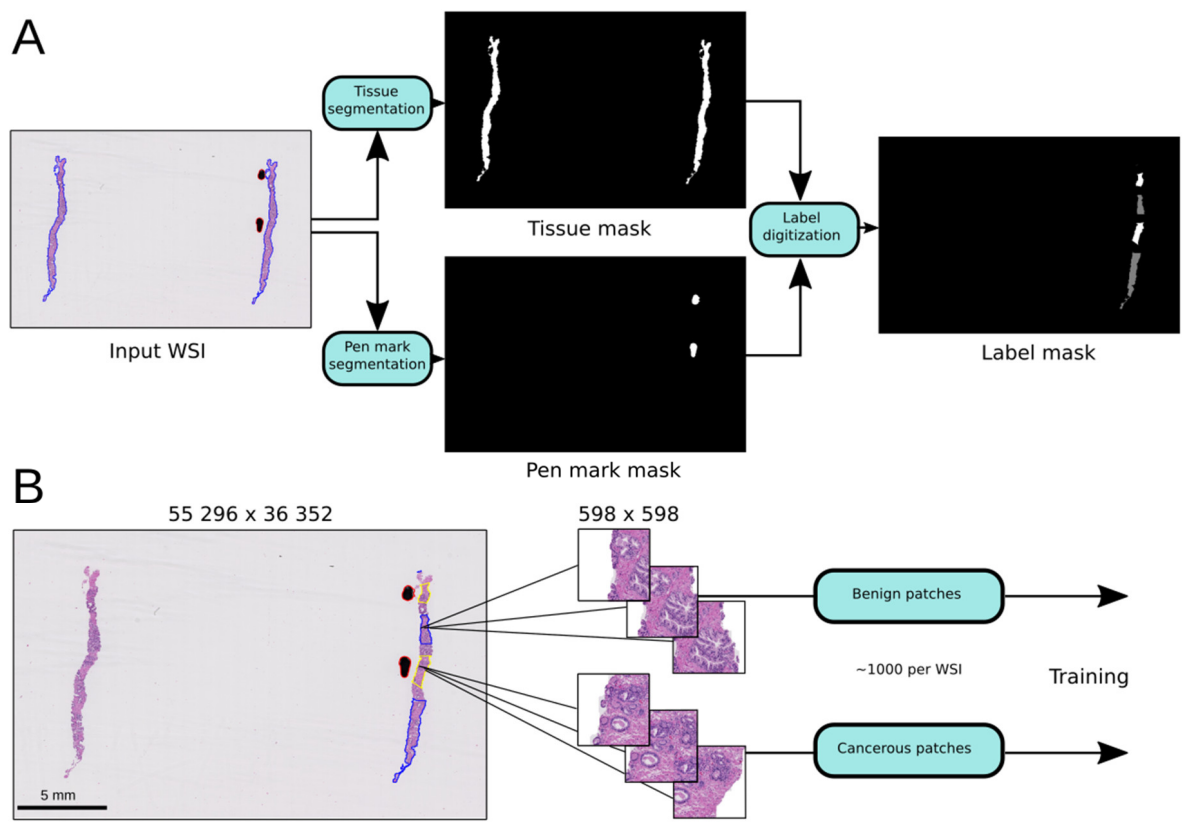

**Figure S1: Image pre-processing workflow. (A)** From left to right: tissue (blue outline) and annotations drawn with a pen (red outline) are segmented from the input WSI and stored as binary masks. The annotations are then digitized by projecting the pen marks onto adjacent tissue, and the result is stored as a label mask indicating benign (grey), cancerous (white) and unknown or background (black) pixels. The unknown label is assigned to non-annotated tissue sections on cancerous slides, such as the one here on the left, and pixels adjacent to cancerous tissue within a specified margin. **(B)** Each WSI is tiled into partially overlapping image patches, extracted from the regions indicated in the tissue mask, resulting in approximately 1000 patches per WSI. In the process, non-tissue pixels are assigned a constant white value to remove any background and pen mark patterns. Each patch is then assigned a label based on the label mask, and benign and cancerous patches are subsequently used as training data. Patches with the unknown label are only used as input in the prediction phase.

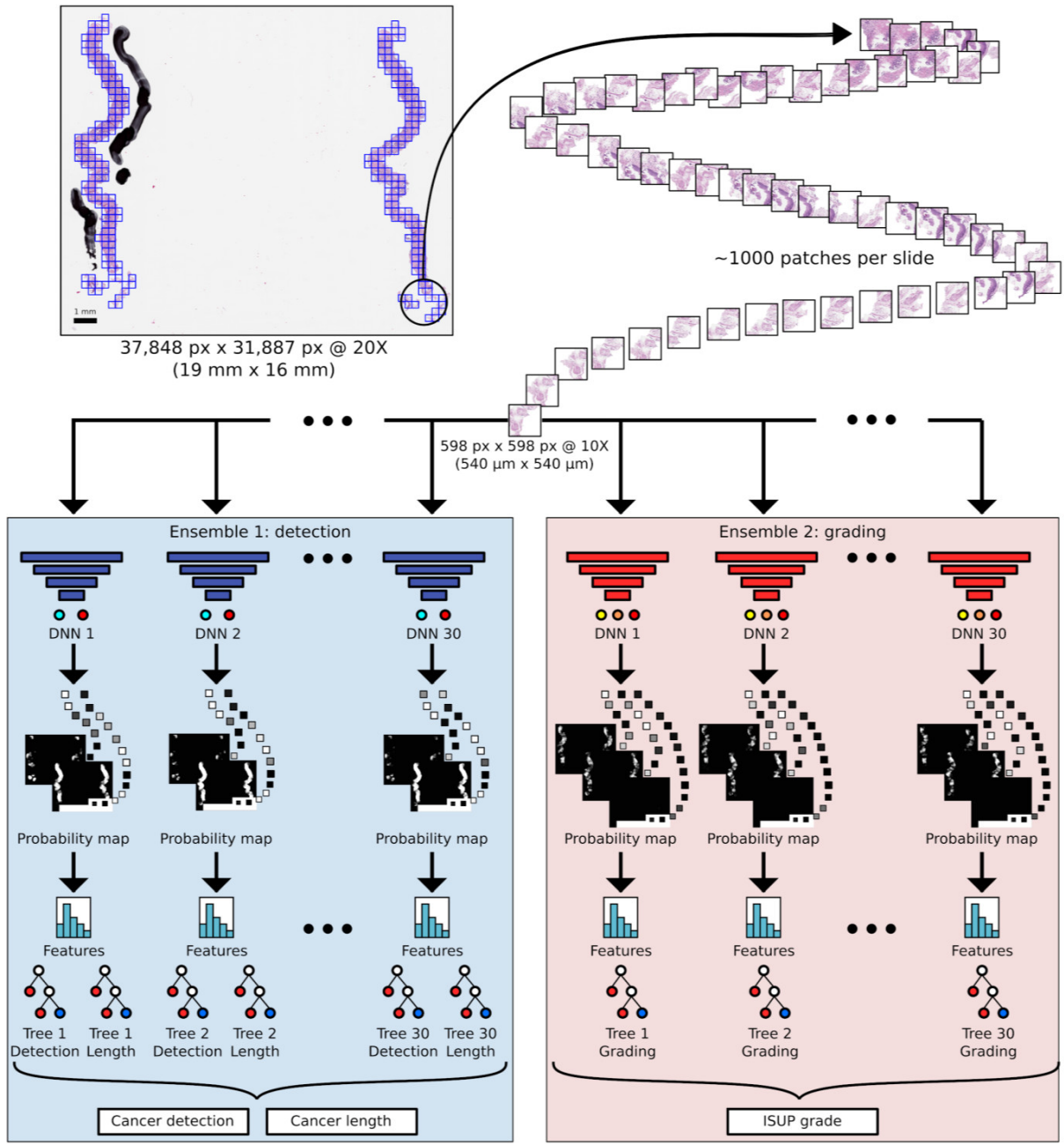

**Figure S2: Overview of the artificial intelligence system.** The tissue region in the input WSI is split into patches **(top)**. The patches are fed as input to a detection ensemble of 30 DNNs for discriminating between benign and malignant patches **(left box; top row)**, and to a grading ensemble of 30 DNNs for classifying patches into Gleason grades 3, 4 and 5 **(right box; top row)**. In the patch-level prediction phase **(both boxes; middle row)**, each trained DNN outputs a vector of class-wise probabilities for each input patch. The class-wise probabilities, indicated here with squares where the grayscale intensity corresponds to probability value, are mapped back to the locations of the corresponding patches in the WSI to construct a probability map for each class. The maps are summarized into features, which are used as inputs to train ensembles of boosted trees to predict cancer presence and extent **(left box; bottom row)** and grade **(right box; bottom row)** for entire WSIs. In the WSI-level prediction phase, outputs from the 30 boosted trees in each ensemble are averaged, and the final classification of each WSI is assigned to the class associated with the highest average probability.

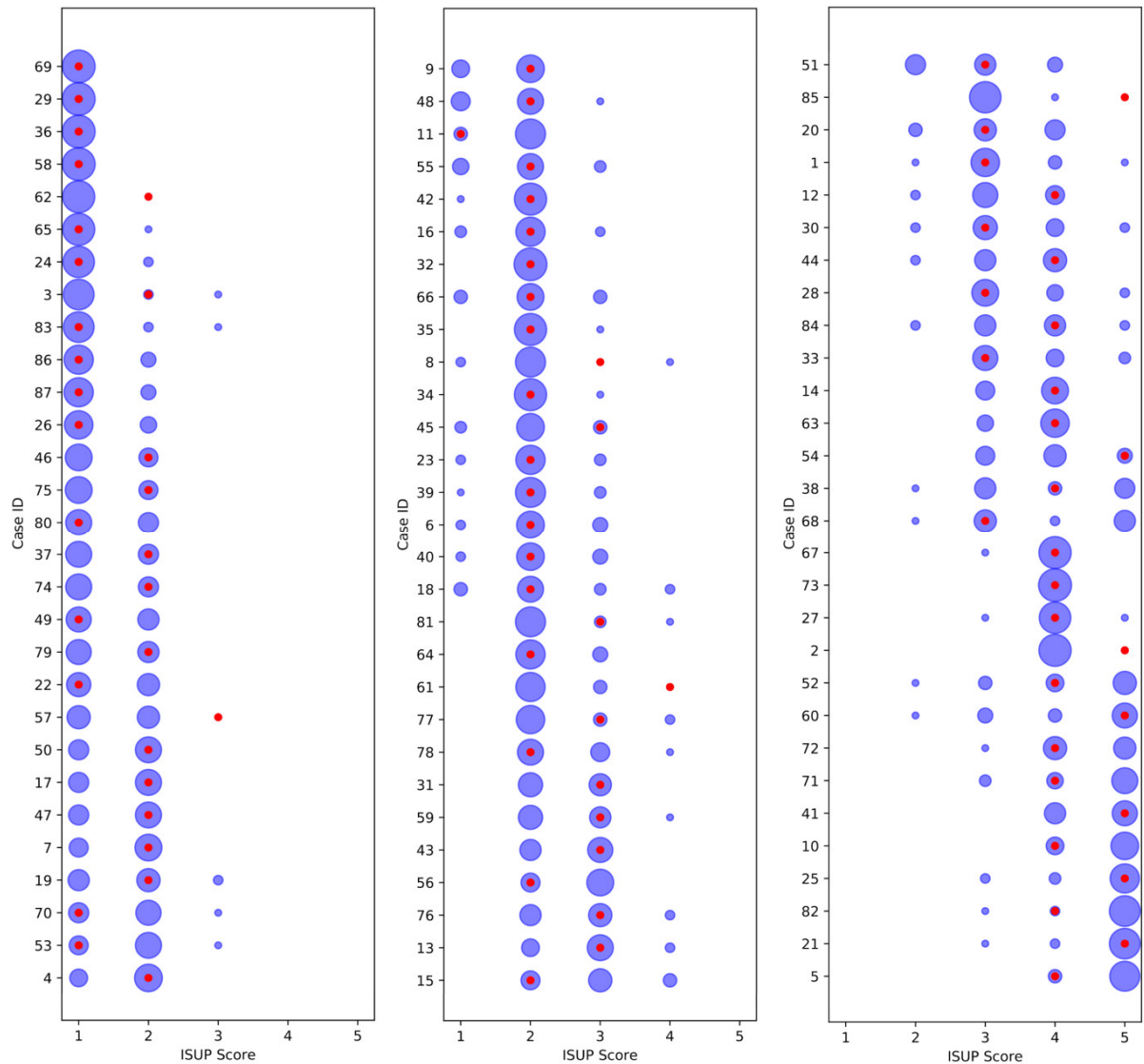

**Figure S3: Grading performance relative to ISUP expert panel on Imagebase.** The distribution of ISUP scores given by the 23 pathologists from the ISUP expert panel and the AI for each of the 87 case IDs in Imagebase. Each row corresponds to one case, and the cases are organized into three plots according to average ISUP score increasing from left to right, and from top to bottom. The areas of the blue circles represent the proportion of pathologists who voted for a specific ISUP score (x-axis). The red dot indicates the ISUP score given by the AI. Example: in the last row (bottom-right; case ID 5) most pathologists voted ISUP 5 and a minority voted ISUP 4; the red dot indicates that the AI voted ISUP 4.

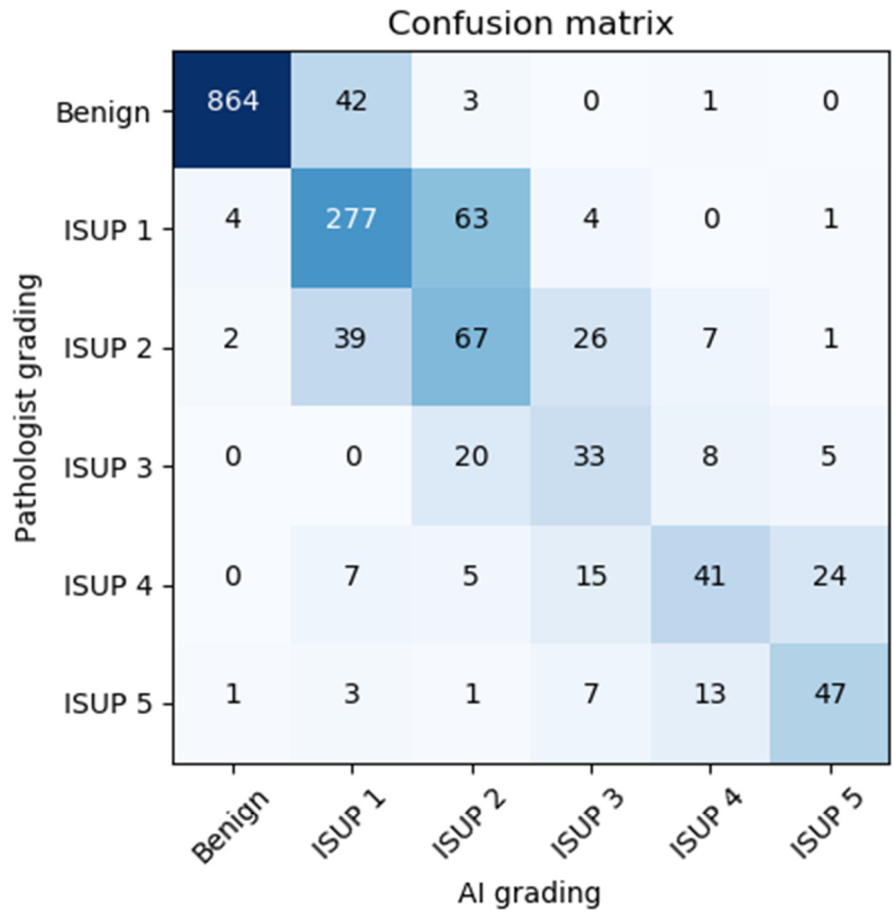

**Figure S4: Grading performance on independent test data.** A confusion matrix on the independent test data of 1631 slides. The pathologist's (L.E.) grading is shown on the y-axis and the AI's grading on the x-axis. Cohen's kappa with linear weights was 0.83 when considering all cases, and 0.70 when only considering the cases indicated as positive by the pathologist.

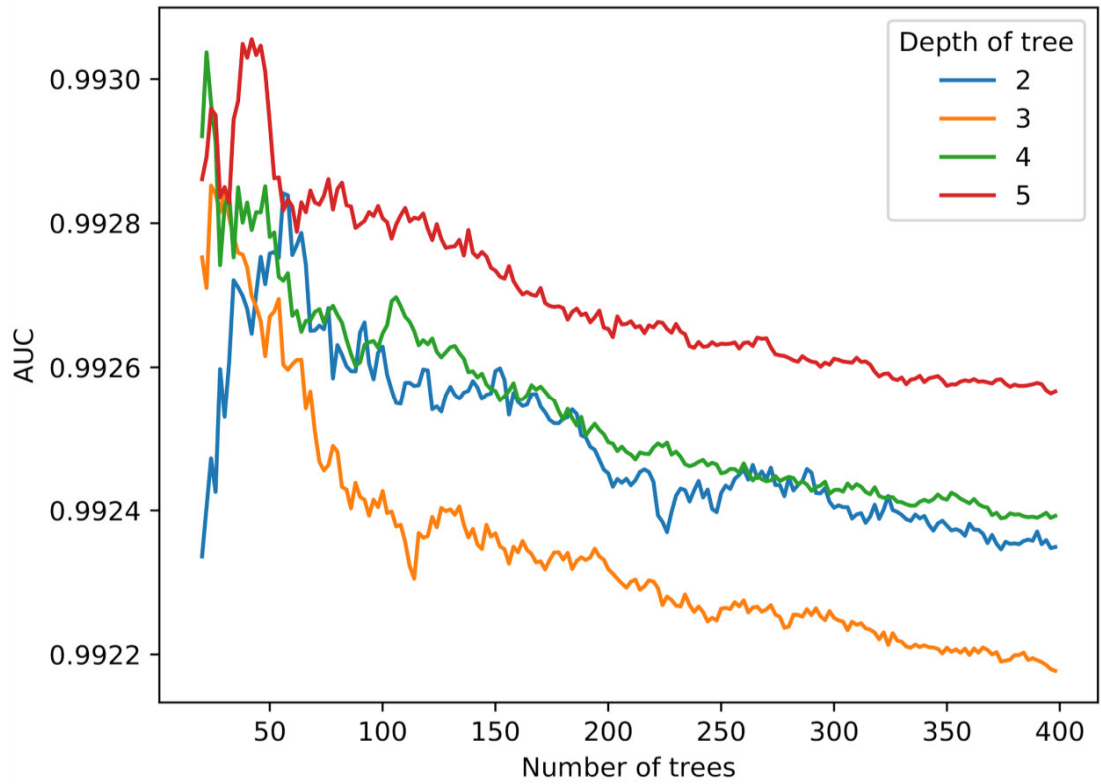

**Figure S5: Effect of boosted tree parameters on cancer detection.** The AUC for discriminating between benign and malignant biopsies is shown for different numbers of trees and tree depths, estimated using cross-validation on the training data.

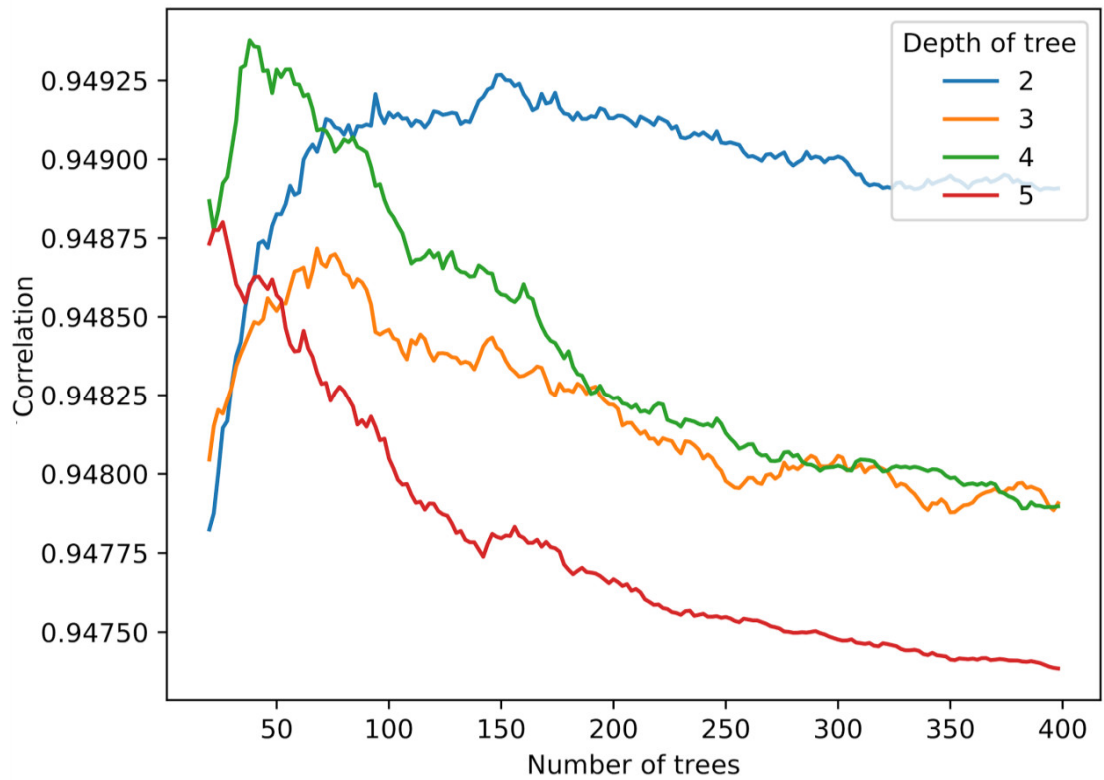

**Figure S6: Effect of boosted tree parameters on cancer length estimation.** The linear correlation coefficient between estimated cancer length and that reported by the study pathologist for each biopsy is shown for different numbers of trees and tree depths, estimated using cross-validation on the training data.

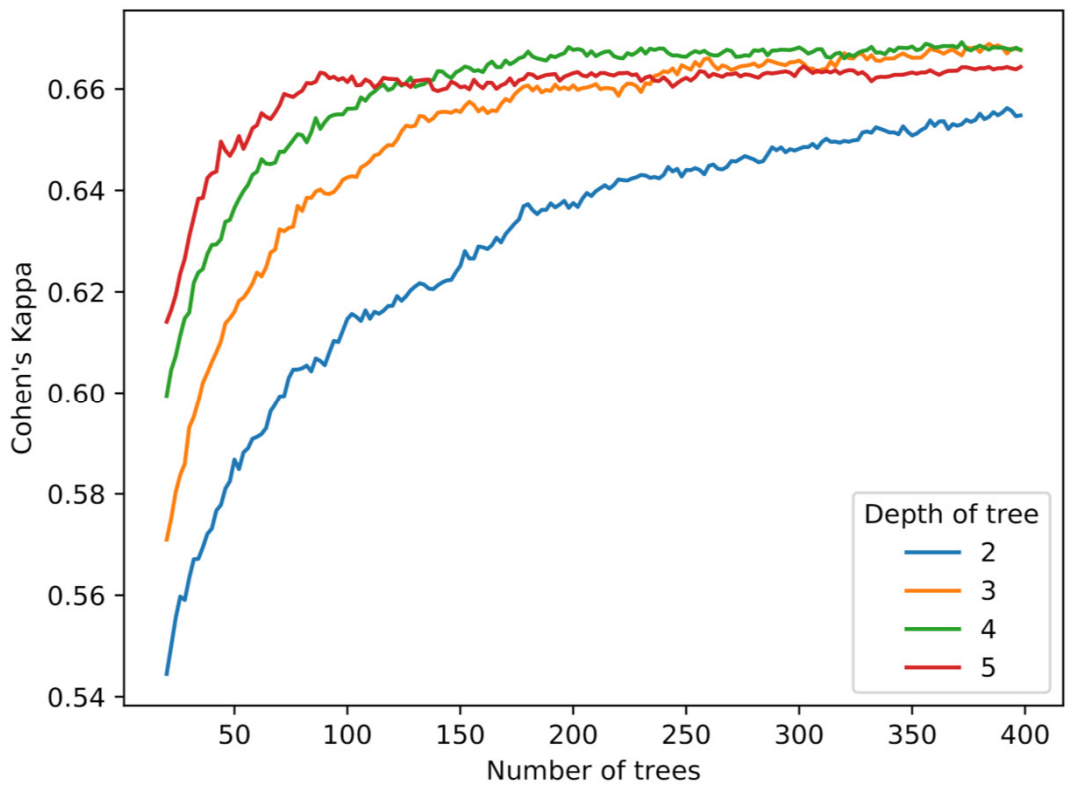

**Figure S7: Effect of boosted tree parameters on cancer grading.** Cohen's kappa between estimated ISUP scores and those reported by the study pathologist for all malignant biopsies is shown for different numbers of trees and tree depths, estimated using cross-validation on the training data.

# 10. Supplementary Tables

**Table S1: Data management workflow and quality control.** Scanned slides were linked back to clinical data. We excluded slides with corrupted filenames, slides that did not pass the visual quality control, slides that were duplicated during scanning and slides that were not consistent with clinical data.

|  | N |
|---|---|
| **Total scanned slides** | 10185 |
| **Excluded slides** |  |
| Non-matching ID filename and clinical data | 268 |
| Rescanned slide | 613 |
| Quality control (IHC, failed staining, corrupted data) | 120 |
| Duplicates scanning | 45 |
| Clinical data and penmark non-matching | 297 |
| **Total remaining slides** | 8842 |

**Table S2: Comparison of CNN architectures.** Cancer detection (AUC), cancer length estimation (Correlation) and grading (Cohen's kappa) performance of ensembles representing different CNN architectures, estimated using cross-validation on the training data. The highest value for each metric is highlighted in bold italic.

| Architecture (epochs) | AUC | Correlation | Cohen's kappa |
|---|---|---|---|
| **Inception V3 (20)** | 0.984 | ***0.943*** | ***0.64*** |
| **Inception V3 (30)** | ***0.987*** | 0.939 | 0.63 |
| **Inception V3 (40)** | 0.984 | 0.935 | 0.62 |
| **ResNet-50 (20)** | 0.983 | 0.933 | 0.57 |
| **ResNet-50 (30)** | 0.980 | 0.935 | 0.60 |
| **ResNet-50 (40)** | 0.982 | 0.932 | 0.55 |
| **Inception-ResNet V2 (20)** | 0.983 | 0.939 | 0.61 |
| **Inception-ResNet V2 (30)** | 0.984 | 0.939 | 0.62 |
| **Xception (20)** | 0.985 | 0.937 | 0.58 |
| **Xception (30)** | 0.985 | 0.939 | 0.58 |

**Table S3: Effect of ensemble size on cancer detection and length estimation performance.** Cancer detection (AUC) and length estimation (Correlation) performance as a function of ensemble size for 1-5 Inception V3 models, estimated using cross-validation on the training data.

|  | n = 1 | n = 2 | n = 3 | n = 4 | n = 5 |
|---|---|---|---|---|---|
| **AUC** | 0.988 | 0.991 | 0.991 | 0.992 | 0.993 |
| **Correlation** | 0.946 | 0.941 | 0.948 | 0.949 | 0.949 |

**Table S4: Effect of training epochs on cancer detection and length estimation performance.** Cancer detection (AUC) and length estimation (Correlation) performance as a function of training epochs for an ensemble of 5 Inception V3 models, estimated using cross-validation on the training data.

|  | 10 epochs | 15 epochs | 20 epochs |
|---|---|---|---|
| **AUC** | 0.991 | 0.993 | 0.992 |
| **Correlation** | 0.947 | 0.949 | 0.948 |

**Table S5: Effect of training epochs on cancer grading performance.** Cancer grading performance (Cohen's kappa) as a function of training epochs for an ensemble of 5 Inception V3 models, estimated using cross-validation on the training data.

|  | 40 epochs | 60 epochs | 80 epochs |
|---|---|---|---|
| **Cohen's kappa** | 0.66 | 0.67 | 0.66 |

**Table S6: Prevalence of different complementary findings.** The number of various histological variants tabulated for the full STHLM3 dataset as well as the training and testing sets used in this study. * Cancer of pseudohyperplastic or atrophic type. § Not including extended training data from StGöran Hospital.

| Complementary findings | All STHLM3 biopsy cores (n=83,470) No. (%) | Training data cores (n=6,682) § No. (%) | Test data cores (n=1,631) No. (%) |
|---|---|---|---|
| **Atypia of unknown significance or suspicious for cancer** | | | |
| No | 82762 (99.2) | 6671 (99.8) | 1627 (99.8) |
| Yes | 708 (0.9) | 11 (0.2) | 4 (0.2) |
| **Prostatic Intraepithelial Neoplasia (PIN)** | | | |
| No | 83424 (99.9) | 6681 (100.0) | 1630 (99.9) |
| Yes | 46 (0.1) | 1 (0.0) | 1 (0.1) |
| **Deceptively bland variants of acinar adenocarcinoma*** | | | |
| No | 83421 (99.9) | 6677 (99.9) | 1627 (99.8) |
| Yes | 49 (0.1) | 5 (0.1) | 4 (0.2) |
| **Adenosis** | | | |
| No | 83458 (100.0) | 6682 (100.0) | 1631 (100.0) |
| Yes | 12 (0.0) | 0 (0.0) | 0 (0.0) |
| **Immunohistochemistry performed** | | | |
| No | 82644 (99.0) | 6569 (98.3) | 1601 (98.2) |
| Yes | 826 (1.0) | 113 (1.7) | 30 (1.8) |

# Supplementary references